\documentclass[12pt,authoryear]{elsarticle}
\usepackage[margin=1in]{geometry}
\usepackage{graphicx}
\usepackage{amsmath}
\usepackage{booktabs}
\usepackage{hyperref}
\usepackage{subcaption}
\usepackage{graphicx}
\usepackage{xcolor}
\usepackage{multirow}
\usepackage{siunitx}
\usepackage[normalem]{ulem}
\usepackage{appendix}
\usepackage{amssymb}
\usepackage{amsmath}
\usepackage[authoryear]{natbib}
\usepackage{float}
\usepackage{placeins}
\journal{}

\begin{document}

\begin{frontmatter}

\title{Bayesian Network Fusion of Large Language Models for Sentiment Analysis}

\author[1]{Rasoul Amirzadeh} 
\author[1]{Dhananjay Thiruvady}
\author[2]{Fatemeh Shiri}
\affiliation[1]{organization={ School of Information Technology, Deakin University},   
            addressline={221 Burwood Highway}, 
            city={Melbourne},
            postcode={3125}, 
            state={Victoria},
            country={Australia}}

\affiliation[2]{organization= {School of Computing Technologies, RMIT University}, 
addressline={124 La Trobe Street}, 
            city={Melbourne},
            postcode={3000}, 
            state={Victoria},
            country={Australia}}

\begin{abstract}
Large language models (LLMs) continue to advance, with an increasing number of domain-specific variants tailored for specialised tasks. However, these models often lack transparency and explainability, can be costly to fine-tune, require substantial prompt engineering, yield inconsistent results across domains, and impose significant adverse environmental impact due to their high computational demands. To address these challenges, we propose the Bayesian network LLM fusion (BNLF)  framework, which integrates predictions from three LLMs, including FinBERT, RoBERTa, and BERTweet, through a probabilistic mechanism for sentiment analysis. 
BNLF performs late fusion by modelling the sentiment predictions from multiple LLMs as probabilistic nodes within a Bayesian network.
 Evaluated across three human-annotated financial corpora with distinct linguistic and contextual characteristics, BNLF demonstrates consistent gains of about six percent in accuracy over the baseline LLMs, underscoring its robustness to dataset variability and the effectiveness of probabilistic fusion for interpretable sentiment classification.
\end{abstract}


\begin{highlights}
\item Proposes a Bayesian Network fusion framework  for integrating multiple LLMs in financial sentiment analysis.
\item Models probabilistic dependencies among LLM predictions to enhance interpretability.
\item BNLF improves accuracy across diverse financial sentiment datasets by up to 6\%.  

\end{highlights}


\end{frontmatter}

\section{Introduction}

Advancements in deep learning, computational power, and large-scale corpora have enabled the development of LLMs. LLMs with billions of parameters, trained through self-supervised learning, achieve state-of-the-art performance in tasks such as text generation, summarization, and sentiment analysis~\citep{raiaan2024review}.
Unlike earlier methods requiring labelled data and manual feature engineering, LLMs leverage prompt-based instructions and pretrained knowledge to classify sentiment with minimal supervision, improving both accuracy and accessibility, especially in zero and few-shot settings~\citep{krugmann2024sentiment}.

Sentiment analysis identifies and interprets opinions in text to assess the author’s attitude toward an entity~\citep{ali2025systematic}. It is typically performed at three levels of document level, sentence level, and aspect level, which focuses on the sentiment toward specific attributes or components mentioned in the text~\citep{wijayanto2018business}. Recent advances in transformer-based LLMs such as BERT and GPT-3 have significantly enhanced sentiment classification by capturing richer contextual information, leading to the development of domain-adapted models fine-tuned on specific data to better reflect the vocabulary and tone of domains ~\citep{mughal2024comparative}. There are various domain-specific models, such as SciBERT~\citep{beltagy2019scibert} for scientific literature and BioBERT~\citep{lee2020biobert} for biomedical and healthcare-related texts.

Although domain-specific LLMs have shown considerable potential, important challenges persist~\citep{pattnayak2025lawmate}. They often underperform compared to general-purpose LLMs in complex tasks, particularly those requiring reasoning or interaction with external tools such as scientific software~\citep{liu2025mattools}. Additionally, domain-specific LLMs are prone to overfitting on the narrow vocabulary and stylistic patterns, which may hinder generalisation and reduce robustness on noisy or heterogeneous inputs~\citep{balaskas2025framework, narashiman2024alphazip}. Another key challenge, particularly in financial applications, is the misalignment between informal, user-generated content and structured reference data. Social media posts, such as tweets, often include sarcasm, slang, or fragmented language, complicating  interpretation compared to  formal texts such as news articles or reports~\citep{kong2024large}.

Beyond these data and structural challenges, scalability and energy efficiency represent major bottlenecks, as modern LLMs demand extensive GPU resources and even domain-specific fine-tuning incurs substantial overhead.\footnote{For example, training costs for GPT-4 were reported at over \$100M, while a single fine-tuning run of BLOOMZ-7B consumed 7571 kW, roughly the annual electricity use of an average U.S. household.} These costs raise concerns about the practicality of deploying LLMs in inference-only or resource-constrained settings~\citep{ gogineni2025llms,xia2024understanding}.
Furthermore, prompt-based interaction with LLMs is often proposed as a lightweight alternative to fine-tuning. However, the performance of domain-specific LLMs remains highly sensitive to prompt phrasing, with minor variations leading to inconsistent outputs~\citep{asadi2025pars}. While fine-tuning can improve adaptation, it introduces trade-offs in as domain-specific gains often reduce generalisation and degrade performance on out-of-domain tasks~\citep{jeong2024fine}.These limitations further emphasise the need for trustworthy AI principles, particularly non-maleficence through resource efficiency and explicability through interpretable models~\citep{thiebes2021trustworthy}.

To address these challenges, we introduce the Bayesian network LLM fusion (BNLF) framework, which uses predictions from three different LLMs, incluing FinBERT, RoBERTa, and BERTweet, to perform sentiment analysis across various curated financial corpora. BNLF leverages the complementary strengths of individual LLMs and dynamically adjusts their influences on sentiment prediction based on the probabilistic dependencies learned through a Bayesian network (BN). While fusion approaches offer a promising direction, the use of BNs for this purpose remains underexplored. BNs are particularly appropriate for this task, as they provide a principled approach for probabilistic reasoning, naturally handle uncertainty and bias, adapt to heterogeneous data sources, and offer interpretable graphical structures for analysing model behaviour~\citep{amirzadeh2023modelling}.

The main contributions of this work are as follows:  
\begin{itemize}
\item We propose a novel framework, called BNLF, that integrates sentiment predictions from multiple LLMs through probabilistic modelling. It is a modular and scalable framework that enhances interpretability and transparent reasoning.  
\item BNLF is designed to be lightweight, leveraging medium-sized LLMs without requiring additional fine-tuning or large-scale GPU resources, making it practical for inference-only sentiment classification tasks.

\item We provide a transparent mechanism to analyse and interpret the behaviour of multiple LLMs  through inference analysis and influence strength analysis.
\item We provide empirical evidence that LLMs trained on particular datasets produce different predictions across data sources, and show how BNLF captures and reveals these effects through probabilistic inference. 

\end{itemize}

The remainder of the paper is structured as follows: Section~\ref{Background} reviews related work, and Section~\ref{methods} outlines the fundamental methods used in the study. Section~\ref{framwork} introduces BNLF framework.  Section~\ref{experiment} describes the experimental setup and data. Section~\ref{Results} presents and discusses the results. Finally, Section~\ref{Conclusions} concludes the paper and outlines future research directions.

\section{Related Work\label{Background}}
Recent studies have systematically benchmarked LLMs against traditional and domain-specific transformers to assess their effectiveness in sentiment analysis.
\citet{krugmann2024sentiment} benchmark GPT-3.5, GPT-4, and Llama 2 against transfer-learning models such as SiEBERT and RoBERTa on 3 900 annotated texts, examining prompting strategies and explainability via human evaluation. Their findings show that LLMs can achieve accurate sentiment classification without fine-tuning but struggle with noisy social-media data, and exhibit systematic biases such as GPT-3.5’s positivity and Llama 2’s tendency to reject offensive content. In another study, \citet{miah2024multimodal} compare GPT-4 and GPT-3.5 with domain-specific transformers (FinBERT, RoBERTa, DistilBERT) across five annotated corpora in finance, healthcare, and social media. The results show that GPT-4 matches fine-tuned models in finance and healthcare, but underperforms on noisy social media data. The study concludes that LLMs provide an alternative to fine-tuning, while still exhibiting biases such as overconfidence and positivity. 

Recent work investigates hybrid strategies that combine domain-specific models, LLMs, and traditional classifiers for sentiment analysis.
\citet{bratic2024centralized} propose a hybrid framework that integrates a transformer-based NLP pipeline with an LLM/chatbot API (GPT-4) to enable centralized access and retrieval of educational materials. The system leverages embeddings, local database search, and interactive dialogue for efficient information access in e-learning environments. Similarly, \citet{miah2024multimodal} propose a multimodal ensemble framework that combines transformer-based models (Twitter-RoBERTa, multilingual BERT) with an LLM (GPT-3) for cross-lingual sentiment analysis. Using datasets in four languages translated into English, the ensemble achieves higher accuracy than individual pre-trained models, demonstrating the benefit of integrating LLMs with traditional transformer architectures for sentiment prediction.
Finally, \citet{tekin2024llm} propose an ensemble framework that integrates outputs from multiple LLMs, such as Llama 2, using diversity-based fusion strategies rather than simple majority voting. Evaluations on multiple reasoning and summarisation benchmarks demonstrate that the ensemble achieves greater robustness and mitigates biases present in single-model predictions.

Building on these ensemble and hybrid approaches, recent research has explored more advanced fusion frameworks within LLM applications. For instance, \citet{das2025presenting} propose FinMSG, an order-aware multimodal transformer architecture that integrates text, audio, and video signals to generate concise financial summaries. The framework introduces a curated dataset (FAV) of over 400 annotated financial expert videos and benchmarks multiple LLMs, including BART, T5, LLaMA-2, and GPT-3.5. Their results highlight the benefit of fusion-based reasoning, achieving around a 7\% improvement. Another fusion approach is MedTsLLM by \citet{chan2024leveraging}, a framework that extends LLM reasoning to time-series analysis in healthcare. The model fuses physiological signals, such as ECG and respiratory data, with textual patient context through a reprogramming layer that performs early feature-level fusion without additional fine-tuning. The framework was evaluated across multiple public medical datasets and achieved performance improvements over existing transformer-based and statistical baselines.

In summary, these studies highlight the growing trend of LLM-centric fusion architectures that integrate heterogeneous data sources to enhance robustness, generalisation, and interpretability. However, most existing approaches remain deterministic, focusing on embedding-level or model-level combinations with limited attention to probabilistic reasoning and dependency modelling, which motivates the design of the BNLF framework.


\section{Preliminaries}\label{methods}
In this section, we briefly introduce the key concepts of LLMs and BNs that form the methodological foundation of this study. For a comprehensive overview, interested readers are refereed to~\citep{minaee2024large} for LLMs and~\citep{polotskaya2024bayesian} for BNs.
\subsection{Large Language Models}

Natural language processing~(NLP) enables computers to interpret and generate human language for tasks such as sentiment analysis, translation, and summarisation. Early NLP approaches relied on rule-based methods and statistical models, including n-grams, which depended heavily on handcrafted features and struggled to capture the complexity of natural language~\citep{muhammad2025similarity}. The emergence of neural networks improved performance through word embeddings and sequential models such as long short-term memory (LSTM) networks, but their sequential nature made training slow and limited their ability to represent long documents effectively~\citep{liu2015fine}.

The breakthrough in transformer architecture replaced recurrent computations with a self-attention mechanism, in which  dependencies among all tokens are computed in parallel to capture contextual relationships across the entire sequence~\citep{shao2024survey}. This mechanism enables efficient modelling of long-range dependencies and scalability to very large datasets, thereby establishing the foundation for subsequent advances in large-scale language modelling. Building on this foundation, LLMs are transformer-based neural networks trained on massive text corpora~\citep{ridley2023enhancing}. With billions or even trillions of parameters, they encode statistical regularities in language and capture semantic and contextual patterns, enabling the model, given an input sequence, to generate the most probable continuation~\citep{chien2024beyond}.

To effectively apply these models, it is necessary to consider how LLMs can be adapted for specific tasks. Two commonly used strategies are fine-tuning and prompt engineering. Fine-tuning is the process in which a pretrained model, such as an LLM, is further trained on task-specific data to adapt it for specialised use cases~\citep{anisuzzaman2025fine}
. The core objective of fine-tuning is to enhance the adaptability of LLMs, enabling them to exhibit behaviour tailored to particular domains or tasks, rather than relying solely on their general pretrained knowledge~\citep{ wu2025llm}.

Prompt engineering, the second strategy, is the process of designing and structuring inputs to guide the output of LLMs for specific tasks. As a relatively new area of research, it focuses on refining prompts or instructions to maximise the utility and accuracy of these models~\citep{mesko2023prompt}. Common prompt engineering approaches include zero-shot and few-shot prompting for in-context learning without retraining, instruction-based and role-defining prompts that frame tasks explicitly, and advanced methods such as chain-of-thought, self-consistency, and generated knowledge prompting to improve reasoning and factual grounding~\citep{mudrik2025prompt, singh2024exploring}. These techniques highlight the adaptability of prompt engineering across domains. However, they also make model outputs highly sensitive to prompt design.

\subsection{Bayesian Networks}

Bayesian networks are a class of probabilistic graphical models designed to represent systems characterised by uncertainty, interdependence, and causal structure~\citep{dewi2025knowledge}. Unlike models that capture only statistical correlations, BNs explicitly encode conditional dependencies between variables, which provide interpretability and make them particularly valuable for reasoning and decision-making in various domains such as finance~\citep{amirzadeh2023framework}, healthcare~\citep{kyrimi2025counterfactual}.

Formally, a BN consists of two main components: a directed acyclic graph (DAG) and a set of conditional probability distributions represented by conditional probability tables (CPTs). In the DAG, nodes correspond to variables in the system, while directed edges denote conditional dependencies from parent (cause) to child (effect). Each node is associated with a CPT that provides the probability of each possible outcome given the states of its parent variables.
The joint probability distribution of all variables in a BN  can be factorised as follows:

$$P(X_1, X_2, \ldots, X_n)=\prod\limits_{i=1}^n P(X_i|\text{parents}(X_i))$$
where $\text{parents}(X_i)$ is the set of parents of $X_i$ in a DAG representation~\citep{heckerman2008tutorial}.

BNs can be constructed using expert knowledge, where domain expertise defines the structure, or through data-driven approaches that automatically learn the DAG and parameters from data. Expert-driven methods may be limited by factors such as subjectivity~\citep{rique2025constructing}, while data-driven methods are often computationally expensive. A hybrid approach integrates both strategies to achieve a balance between interpretability and accuracy~\citep{polotskaya2024bayesian}.

\section{Proposed Framework: Bayesian Network  LLMs Fusion 
}\label{framwork}

\begin{figure}[h!]
\centering
\includegraphics[width=\textwidth, keepaspectratio]{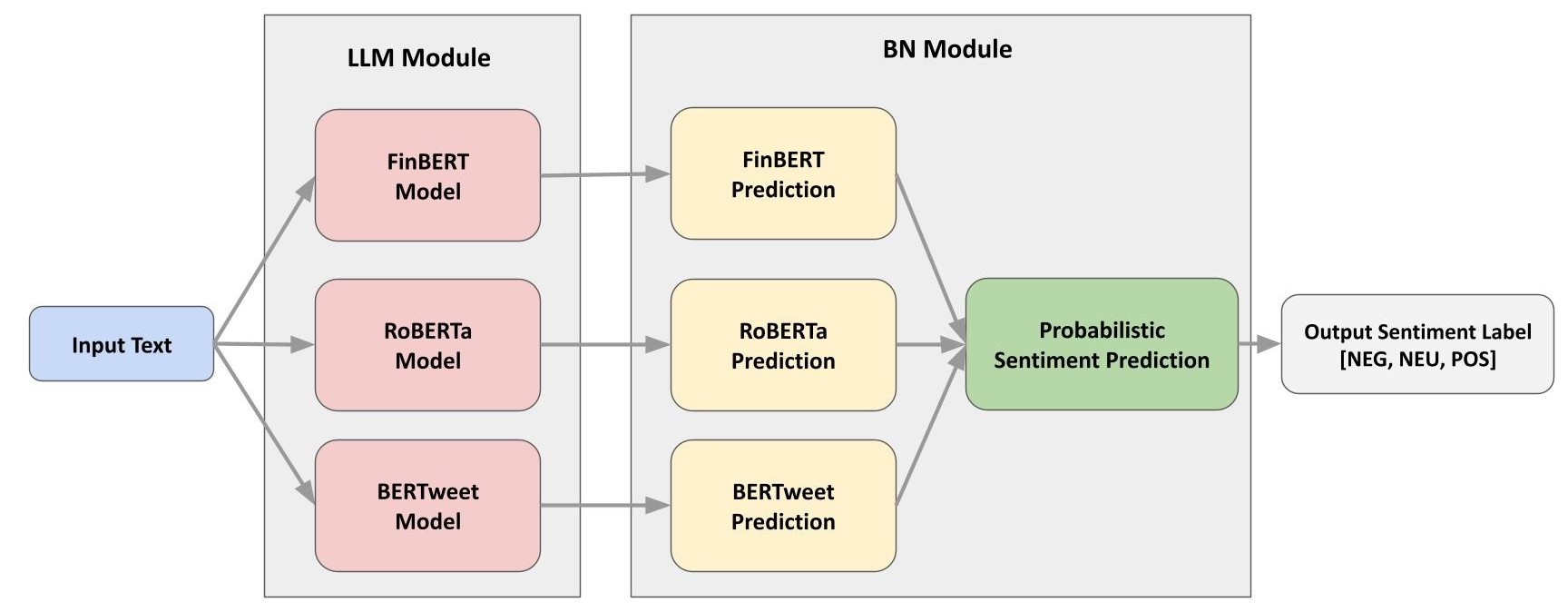}
\caption{Each input text, drawn from multiple financial and social media corpora, is processed by three LLM-based classifiers: FinBERT, RoBERTa, and BERTweet. These models generate individual sentiment predictions that are fused within a BN through probabilistic inference. The BN outputs a posterior sentiment distribution, which is then mapped to a discrete sentiment label: negative (NEG), neutral (NEU), or positive (POS).
}
\label{fig:bnlf_structure}
\end{figure}

In this section, we introduce the Bayesian network LLM Fusion (BNLF) framework as an end-to-end pipeline for sentiment analysis, as illustrated in Figure~\ref{fig:bnlf_structure}. The framework has four components. First, the texts are drawn from various corpus sources. Second, these texts are processed by three LLMs (FinBERT, RoBERTa, and BERTweet), each producing sentiment predictions. Third, predictions are fused through a BN to produce a probabilistic sentiment inference. Finally, the fused inference produces a discrete sentiment label. Thus, BNLF represents the entire pipeline from input text to final classification, while BN specifically constitutes the probabilistic fusion mechanism within it.

To illustrate the role of each component, we begin with the inputs. Each \texttt{Input Text}, drawn from multiple financial corpora, serves as the entry point to the BNLF pipeline. The corpora  span both formal financial documents and informal user-generated content. Financial text poses unique challenges for sentiment analysis because it differs from natural conversational language and often combines technical vocabulary with alphanumeric expressions and context-dependent cues. This diversity  introduces systematic variation in the sentiment distribution of financial texts, which the \texttt{Input Text} is designed to capture as a contextual factor within the framework.

In addition to the dataset-level signal, the framework includes an \texttt{LLM Module} that processes input texts to generate sentiment predictions. As shown in Figure~\ref{fig:bnlf_structure}, three models of RoBERTa, BERTweet, and FinBERT are employed. RoBERTa is a general purpose transformer pretrained on a large and diverse corpus and later fine-tuned for sentiment tasks, including Twitter-based datasets~\citep{mohawesh2024fake}. BERTweet follows a similar pretraining procedure but is  trained on a large-scale Twitter data, enabling it to capture the informal and noisy language characteristic of social media~\citep{baker2022classification}. FinBERT adapts the BERT architecture by training on both general text and financial documents, making it sensitive to finance-specific sentiment cues~\citep{karadacs2025multimodal}. Together, these  models capture complementary aspects of sentiment across financial and social media corpora, forming the predictive foundation for BNLF.

These models are chosen for their complementary coverage, efficiency, and practicality. FinBERT provides domain-specific knowledge of the financial language, RoBERTa serves as a strong general-purpose model, and BERTweet captures the informal style of user-generated content. A key design goal of BNLF is to remain lightweight while robust, enabling efficient inference without the high GPU costs of larger models. Each of the three models is medium-sized ($\approx$110M–135M parameters)\footnote{For comparison, FinLLaMA variants range from 7B to 13B parameters, requiring 13–24 GB of storage in full precision and high-end GPUs ($\ge$ 24 GB VRAM) for fine-tuning~\citep{chen2025inference}. Similarly, models such as FinSoSent have been reported in the literature but remain without publicly available pretrained weights, making them unsuitable for inference-only setups.}, allowing deployment on CPUs or single-GPU systems. BNLF thus combines computational efficiency with practical robustness.

Next, after defining the inputs  and the LLMs, we specify the fusion mechanism, where predictions from the LLMs are combined using a BN. Fusion strategies are generally classified as early (data-level), deep (feature-level), late (decision-level), and hybrid fusion~\citep{yi2025survey}. BNLF employs a late fusion strategy, where sentiment predictions from individual LLMs are fused using a BN that integrates multiple predictions into a  probabilistic estimate. Unlike standard ensemble methods that aggregate predictions via majority voting or averaging, the BN explicitly models the joint probability distribution over the input variables, enabling a principled probabilistic fusion that captures uncertainty in the final sentiment inference.\footnote{In preliminary experiments, we also evaluated fuzzy integration of confidence scores, which yielded suboptimal results. Such baselines collapse multiple signals into a single score and fail to capture dataset-conditioned dependencies or offer an interpretable joint model, motivating the BN-based fusion.}

Having established the role of BNLF as a late-fusion pipeline, we now describe the internal structure of its BN component in more detail. The BN fuses LLM predictions through probabilistic inference, while also capturing how dataset context and individual model outputs jointly influence the final sentiment decision. To achieve this, the BN is represented by a DAG, where arcs encode conditional relationships among the \texttt{Input Text}, the LLM prediction nodes, and the \texttt{Probabilistic Sentiment Prediction} node.

In the  DAG, the \texttt{Input Text} is connected to each LLM prediction node to model systematic domain effects on model behaviour, and the prediction nodes are connected to the \texttt{Probabilistic Sentiment Prediction} node so  their outputs directly inform sentiment inference. No arcs are allowed between the LLM prediction nodes, enforcing conditional independence and preventing correlations among them from biasing results. This structure ensures interpretability, reflects domain-specific relationships, and provides a principled basis for probabilistic fusion. Based on the DAG specified from domain knowledge, CPTs are subsequently learned from the annotated training data.\footnote{It should be noted that   DAG described is distinct from the pipeline architecture of BNLF shown in Figure~\ref{fig:bnlf_structure}. The BN specifies probabilistic dependencies among variables, while the framework figure illustrates the overall flow from input text to final sentiment classification. The two views are complementary but not equivalent.
}

\begin{figure}[h!]
    \centering
    \includegraphics[width=\textwidth, keepaspectratio]{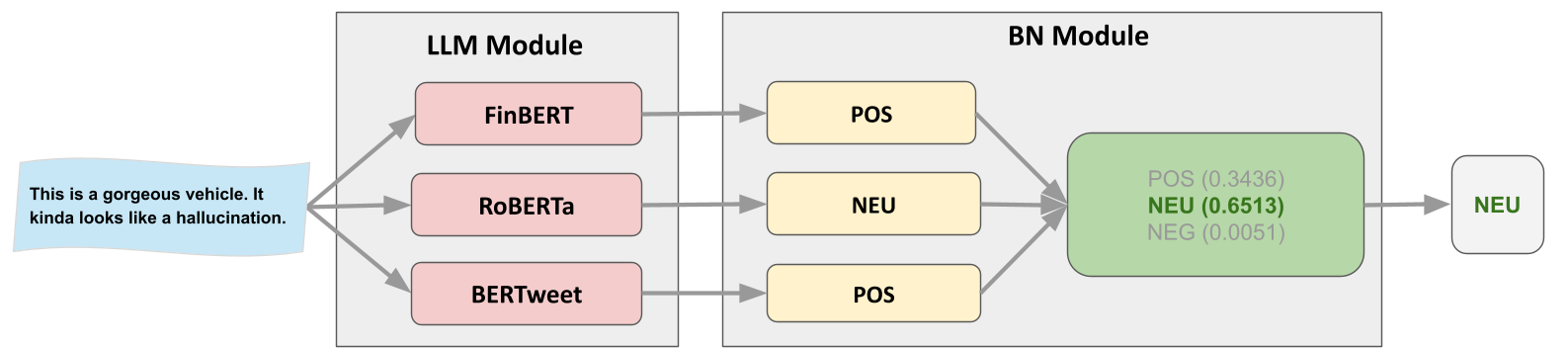}
    \caption{A real example from the dataset showing how BNLF fuses sentiment predictions from FinBERT, RoBERTa, and BERTweet. BN integrates these individual predictions to generate final sentiment probabilities of (POS = 0.3436, NEU = 0.6513, NEG = 0.0051), where the neutral class is eventually selected with the highest probability.}
    \label{fig:bnlf_example}
\end{figure}

To provide a clearer understanding of how BNLF performs, Figure \ref{fig:bnlf_example} presents a real example from the dataset, showing how sentiment predictions from three LLMs are fused through the BN to produce a final probabilistic sentiment output.

\section{Datasets and Experimental Design \label{experiment}}

To assess the proposed framework, we require diverse, annotated datasets that capture different styles of financial text. To this end, we evaluate BNLF using three publicly available, human-annotated corpora that provide gold-standard sentiment labels (positive, neutral, negative) assigned by expert or crowd-sourced annotators, as follows.

\begin{itemize}
\item\textbf{Financial PhraseBank:} A news-based dataset for financial sentiment classification, annotated by professional financial analysts. It consists of 4,840 short text fragments drawn from corporate announcements, market news, and financial reports. Since the data originates from authentic financial news sources, it reflects policy discourse and risk-related language. Annotations were provided by 5-8 experts, and the dataset is also divided based on the agreement rate among annotators~\citep{Malo2014GoodDO}.

\item \textbf{Twitter Financial News Sentiment~(TFNS):} An English-language dataset of 12,424 finance-related tweets annotated with three sentiment classes: bearish, neutral, and bullish. This dataset reflects the informal, concise, and time-sensitive nature of financial discourse on Twitter where human annotations ensure reliable categorisation of sentiment~\citep{zeroshot2024twitterfinnews}.

\item \textbf{FIQA:} 
Originally developed for the financial opinion mining and question answering challenge (FIQA-2018), this dataset contains 961 entries covering diverse financial topics, including market events, company performance, and investment opinions. It is included in the BEIR benchmark, which standardises preprocessing and format across multiple retrieval and sentiment datasets. Each entry is labelled as positive, neutral, or negative~\citep{thakur2021beir}.
\end{itemize}

All three datasets are publicly available on Hugging Face~\citep{wolf2019huggingface}
, ensuring transparent and reproducible experimentation. They span formal corporate disclosures to informal social media posts, enabling assessment across diverse linguistic and contextual settings. Their combination supports robust LLM evaluation and reliable BN parameter learning.

Before experimentation, the datasets were merged into a single corpus and underwent essential preprocessing, including removal of missing or empty texts, deletion of zero-width characters, spacing normalisation, and duplication removal. 
 Sentiment labels were remapped to a consistent three-class scheme (0 = negative, 1 = neutral, 2 = positive) to ensure comparability across datasets. For instance, in the {TFNS} dataset the original classes bearish, bullish, and neutral were mapped to negative, positive, and neutral, respectively. {Table~\ref{tab:dataset_stats} summarises the class distributions for each dataset.

\begin{table}[h!]
\centering
\caption{Summary of sentiment dataset statistics showing the number and proportion of negative, neutral, and positive samples in each corpus.}
\begin{tabular}{lrrr}
\hline
\textbf{Dataset} & \textbf{Negative} & \textbf{Neutral} & \textbf{Positive} \\
\hline

Financial PhraseBank & 303 (13.38\%) & 1{,}391 (61.44\%) & 570 (25.18\%) \\
TFNS & 1{,}789 (14.99\%) & 7{,}744 (64.91\%) & 2{,}398 (20.10\%) \\
FIQA & 716 (59.03\%) & 118 (9.73\%) & 379 (31.24\%) \\
\hline
\textbf{Total 
} & {2{,}808 (18.22\%)} & {9{,}253 (60.05\%)} & {3{,}347 (21.72\%)} \\
\hline
\end{tabular}
\label{tab:dataset_stats}
\end{table}

This unified dataset was then split into training and test partitions using an 80/20 ratio, a common practice in BN studies, applied consistently for both BN construction and LLM evaluation. Following the BNLF design, the BN structure was constrained by domain knowledge and its CPT parameters were learned from the training set. To ensure comparability, the BNLF model was evaluated on the test set, which was also used for standalone LLM evaluation.

Alongside the three models used within BNLF, we also evaluate the performance of DistilRoBERTa-financial-sentiment
as an external baseline~\citep{distilroberta_financial}. This model serves as an independent benchmark due to its fine-tuning on financial data, computationally efficient (82M parameters), and is widely adopted in Hugging Face. Its popularity, high performance in financial sentiment tasks~\citep{kadasi2025model}, and adoption in various recent studies~\citep{banka2025options} ensure both robustness and practical accessibility. Thus, BNLF is compared both against its constituent LLMs and with an external benchmark.\footnote{We also tested additional models, but their results were substantially weaker. The complete results are reported in~\ref{sup_res} for completeness.}

In addition to the individual LLMs and the external DistilRoBERTa baseline, two ensemble baselines were included for comparison. The first, majority voting, assigns each instance the sentiment label predicted by at least two of the three constituent LLMs—FinBERT, RoBERTa, and BERTweet. In cases where no majority exists, FinBERT’s prediction is used, reflecting its financial-domain specialisation. The second, averaging, aggregates class probabilities across  three LLMs and selects the sentiment with the highest mean probability.

The model’s performance is evaluated using multiple complementary metrics. Accuracy  captures overall correctness, while macro- and weighted F1-scores provided balanced assessments of performance across sentiment classes. In addition, model agreement is assessed through a pairwise agreement matrix and Cohen’s Kappa, which analyse consensus and divergence across models.

Experiments were conducted on a Windows 10 Enterprise system equipped with an Intel Core i7 CPU and 16 GB of RAM. All implementations were developed in Python. LLM experiments were carried out with the Hugging Face Transformers library.
The three LLMs were applied through sentiment-analysis pipelines in inference-only mode, using fixed pretrained weights to generate sentiment predictions without task-specific fine-tuning. This setup isolates model comparison from fine-tuning effects and evaluates the inherent generalisability of pretrained LLMs across financial corpora.

BNs were implemented and evaluated using GeNIe and pySMILE~\citep{druzdzel1999smile}. To construct the structure according to the proposed BNLF design, we employed GeNIe’s \textit{Knowledge Editor} to incorporate domain-specific background knowledge. The default hyperparameter settings for the BN are reported in Table \ref{tab:genie_config}.

\begin{table}[!ht]
\centering
\caption{
Hyper-parameters used for BN construction.}
\begin{tabular}{ll}
\textbf{Parameter}           & \textbf{Value} \\
\hline
Learning Algorithm           & Bayesian Search \\
Discrete Threshold           & 20 \\
Max Parent Count             & 8 \\
Iterations                   & 20 \\
Sample Size                  & 50 \\
Seed                         & 0 \\
Link Probability             & 0.1 \\
Prior Link Probability       & 0.001 \\
\hline
\end{tabular}%

\label{tab:genie_config}
\end{table}

\section{Results and Discussion}\label{Results}

This section summarises key experimental results and comparative analyses of BNLF and baseline models.
Table~\ref{tab:overall_results} presents the results in a zero-shot setting across three metrics for the combined test set derived from all three sources. It shows that BNLF 
achieves the best results, with 78.6\% accuracy, about 5.3\% higher than the DistilRoBERTa baseline. It also improves both macro- and weighted-F1 scores, indicating that the improvement is consistent across classes. Furthermore, DistilRoBERTa outperforms all of the individual LLMs included in BNLF, underscoring its strength as a competitive baseline. In addition, among the three models that form the basis of BNLF, FinBERT is the best performer (72.1\% accuracy), reflecting its financial-domain specialisation. RoBERTa and BERTweet achieve comparable accuracies (65.8\% and 66.2\%, respectively) but lower F1-scores, which is consistent with their shared transformer architecture and broader training domains.


The ensemble baselines achieve moderate performance, with majority voting attaining 72.6\% accuracy and probability averaging reaching 74.8\%. Both results are less than BNLF’s 78.6\%, suggesting that simple aggregation methods offer limited improvement over individual models. In contrast, BNLF delivers more consistent and substantial gains across all metrics. Compared with traditional ensemble methods, BNLF provides a more principled integration of multiple LLMs. Instead of combining outputs linearly, it models conditional dependencies among LLMs through a probabilistic structure. This results in a more informed fusion process and greater robustness.
For better comparison, we also visualise the overall performance metrics in Figure~\ref{fig:overall_performance}.

\begin{table}[h!]
\centering
\caption{Performance comparison of individual LLMs, ensemble baselines, and BNLF on the combined test set. DistilRoBERTa serves as the external baseline, while majority voting and averaging are ensemble baselines combining FinBERT, RoBERTa, and BERTweet. Best results are shown in bold.}
\begin{tabular}{lccc}
\hline
\textbf{Model} & \textbf{Accuracy (\%)} & \textbf{Macro-F1 (\%)} & \textbf{Weighted-F1 (\%)} \\
\hline
DistilRoBERTa (baseline) & 73.3 & 68.3 & 73.8 \\
RoBERTa                  & 65.8 & 54.8 & 63.6 \\
BERTweet                 & 66.2 & 57.4 & 65.0 \\
FinBERT                  & 72.1 & 66.7 & 72.5 \\
Majority Voting & 72.6 & 65.1 & 71.9 \\
Averaging        & 74.8 & 68.5 & 74.4 \\
{BNLF}            & \textbf{78.6} & \textbf{71.5} & \textbf{77.7} \\
\hline
\end{tabular}
\label{tab:overall_results}
\end{table}

\begin{figure}[!h]
    \centering
\includegraphics[width=\linewidth, keepaspectratio]{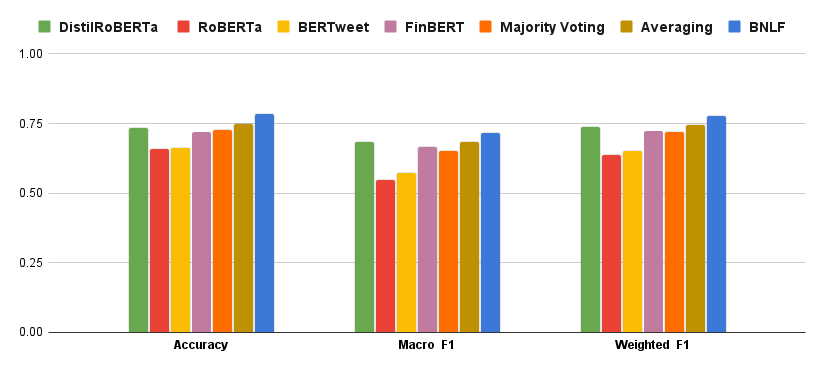}       \caption{Overall performance comparison of BNLF, individual LLMs, ensemble baselines, and the external DistilRoBERTa model across accuracy, macro-F1, and weighted-F1 metrics. The blue bars representing BNLF consistently exceed those of all baselines, including the ensemble methods (majority voting and averaging), demonstrating the effectiveness of its probabilistic fusion approach.} 

\label{fig:overall_performance}
\end{figure}

Class-level analysis is important for assessing whether a model’s performance is balanced across majority and minority classes.
To this end, Table~\ref{tab:class_metrics} reports precision, recall, and F1-scores for each sentiment class. As can be seen in the table, BNLF achieves balanced performance across classes, with the highest scores in neutral sentiment (F1 = 0.850, recall = 0.912) and competitive results in negative (F1 = 0.639) and positive (F1 = 0.667) classes.  Overall, DistilRoBERTa performs better across most classes than the other individual LLMs, but its performance remains lower than that of BNLF. While it is strong on the neutral class, its results on the negative and positive categories are weaker compared to BNLF. 

\begin{table}[h]
\centering
\caption{Class-level 
precision, recall, and F1-score for LLM models and BNLF. 
Classes correspond to sentiment categories, including negative, neutral, and positive. 
Best values in each column are shown in bold.
}
\resizebox{\textwidth}{!}{%
\begin{tabular}{lccccccccc}
\hline
 & \multicolumn{3}{c}{Negative} & \multicolumn{3}{c}{Neutral} & \multicolumn{3}{c}{Positive} \\
Model & Precision  & Recall  & F1 & Precision  & Recall  & F1 & Precision  & Recall  & F1 \\
\hline
DistilRoBERTa (baseline)      & 0.576 & 0.596 & 0.586 & \textbf{0.860} & 0.789 & 0.823 & 0.595 & \textbf{0.715} & 0.649 \\

RoBERTa                       & 0.525 & 0.353 & 0.422 & 0.692 & 0.850 & 0.763 & 0.585 & 0.375 & 0.457 \\
BERTweet                      & 0.515 & 0.452 & 0.482 & 0.707 & 0.797 & 0.749 & 0.537 & 0.402 & 0.460 \\
FinBERT                       & 0.500 & \textbf{0.643} & 0.563 & 0.840 & 0.762 & 0.799 & 0.620 & 0.631 & 0.626 \\
BNLF   & \textbf{0.684} & 0.600 & \textbf{0.639} & 0.796 & \textbf{0.912} & \textbf{0.850} & \textbf{0.848} & 0.548 & \textbf{0.667} \\     
\hline
\end{tabular}
}
\label{tab:class_metrics}
\end{table}

While overall results provide a high-level view of performance, a closer examination across datasets further reveals how the models adapt to varying linguistic styles and domain characteristics.
To assess BNLF’s performance across different datasets, Table~\ref{tab:per_source_general} reports the results for each data source individually. Our framework outperforms all LLMs on FIQA and TFNS, while DistilRoBERTa achieves slightly higher accuracy on {Financial PhraseBank}. In particular, on {Financial PhraseBank}, DistilRoBERTa achieves the highest accuracy (99.6\%), while BNLF also performs strongly with 98.2\% accuracy and balanced F1-scores, surpassing FinBERT’s already strong results. For {FIQA}, which contains more diverse and nuanced financial question–answer content, BNLF improves accuracy to 65.8\%, higher accuracy over the best individual model (BERTweet at 60.1\%). On {TFNS}, the fusion approach reaches 75.3\% accuracy, outperforming FinBERT’s 73.2\% and showing consistent improvements in both macro- and weighted-F1.  

\begin{table}[H]
\centering
\caption{Comparison of model performance across individual datasets (Financial PhraseBank, FIQA, and TFNS). Results are reported with accuracy, macro-F1, and weighted-F1, with best values highlighted in bold.
 }
\begin{tabular}{c l S[table-format=1.4] S[table-format=1.4] S[table-format=1.4]}
\hline
\textbf{Source} & \textbf{Model} & \textbf{Acc} & \textbf{Macro-F1} & \textbf{Weighted-F1} \\
\hline
\multirow{5}{*}{Financial PhraseBank} 
 & DistilRoBERTa~(baseline) & \textbf{0.9956} &   \textbf{0.9919}   &   \textbf{0.9956} \\
 & RoBERTa     & 0.7035 & 0.5182 & 0.6463 \\
 & BERTweet    & 0.7788 & 0.6873 & 0.7565 \\
  & FinBERT     & 0.9690 & 0.9593 & 0.9694 \\
 & BNLF        & 0.9823 & 0.9811 & 0.9820 \\
\hline
\multirow{4}{*}{FIQA} 
  &DistilRoBERTa~(baseline)& 0.4571  &      0.4179   &        0.5393\\

 & RoBERTa     & 0.5247 & 0.3823 & 0.5442 \\
 & BERTweet    & 0.6009 & 0.4581 & 0.6374 \\
  & FinBERT     & 0.5561 & 0.4491 & 0.6114 \\
 & BNLF        & \textbf{0.6583} & \textbf{0.5224} & \textbf{0.6586} \\
\hline
\multirow{4}{*}{TFNS} 
&DistilRoBERTa~(baseline)  &  0.7452   & 0.7015&       0.7529 \\

 & RoBERTa     & 0.7030 & 0.6101 & 0.6947 \\
 & BERTweet    & 0.6921 & 0.6153 & 0.6877 \\
  & FinBERT     & 0.7323 & 0.6794 & 0.7400 \\
 & BNLF        & \textbf{0.7525} & \textbf{0.6997} & \textbf{0.7501} \\
\hline
\end{tabular}
\label{tab:per_source_general}
\end{table}

To visualise dataset-level performance, we include Figure~\ref{fig:acc_per_dataset}, which compares the accuracy of all models based on the datasets.  

\begin{figure}[H]
  \centering
\includegraphics[width=0.85\linewidth, keepaspectratio]{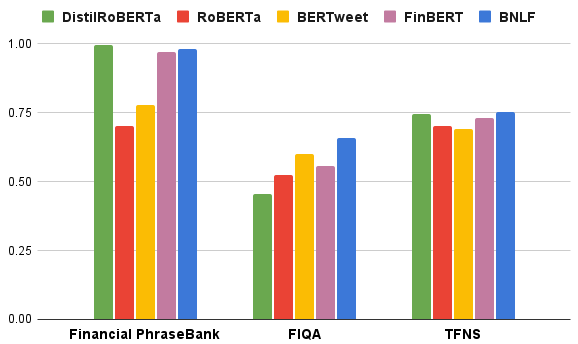} 
  \caption{Accuracy comparison across datasets, where each group of bars corresponds to one dataset (Financial PhraseBank, FIQA, TFNS), with  bars representing indivicual LLMs, and BNLF. It shows that BNLF achieves the highest accuracy on FIQA and TFNS, while DistilRoBERTa reaches the highest accuracy on Financial PhraseBank.}
  \label{fig:acc_per_dataset}
\end{figure}

\subsection{Model Agreement Analysis} 

Pairwise agreement scores quantify how closely two models align in their predictions, offering a straightforward measure of similarity in classification behaviour. To this end, this section compares the prediction patterns of individual LLMs and BNLF by measuring the proportion of documents assigned identical sentiment labels. We also report Cohen’s Kappa, which adjusts for chance agreement and provides a more robust measure of alignment. The results are presented in Table~\ref{tab:agreement_combined}.

\begin{table}[h!]
\centering
\caption{Pairwise model agreement in which each cell shows the proportion of matching sentiment labels. Cohen’s Kappa is presented in parentheses. The “Mean”  column reports the average agreement for each model.}
\label{tab:agreement_combined}
\resizebox{\textwidth}{!}{%
\begin{tabular}{lccccc|c}
\toprule
\textbf{Model} & \textbf{DistilRoBERTa} & \textbf{FinBERT} & \textbf{RoBERTa} & \textbf{BERTweet} & \textbf{BNLF} & \textbf{Mean} \\
\midrule
DistilRoBERTa  & 1.000 & 0.783 (0.648) & 0.671 (0.423) & 0.674 (0.440) & 0.736 (0.543) & 0.716 \\
FinBERT        &       & 1.000 & 0.659 (0.396) & 0.682 (0.448) & 0.831 (0.704) & 0.739 \\
RoBERTa        &       &       & 1.000 & 0.824 (0.643) & 0.799 (0.580) & 0.738 \\
BERTweet       &       &       &       & 1.000 & 0.838 (0.677) & 0.754 \\
BNLF           &       &       &       &       & 1.000 & 0.801 \\
\bottomrule
\end{tabular}
}
\end{table}

As can be seen in the table, DistilRoBERTa shows strong alignment with FinBERT (78.3\%) and BNLF (73.6\%), while RoBERTa and BERTweet demonstrate the closest relationship (82.4\%), reflecting their shared architecture and training domains. BNLF exhibits consistently high agreement with all three transformer-based models, suggesting that the fusion effectively integrates their complementary patterns. On average, BERTweet (75.4\%) and BNLF achieves the highest mean consistency (80.1\% based on proportions), whereas DistilRoBERTa (71.6\%) is slightly less aligned with the other models.
Furthermore, the results are shown in Figure~\ref{fig:agreement_heatmap} for better comparison of the models.

\begin{figure}[!h]
    \centering
\includegraphics[width=\linewidth, keepaspectratio]{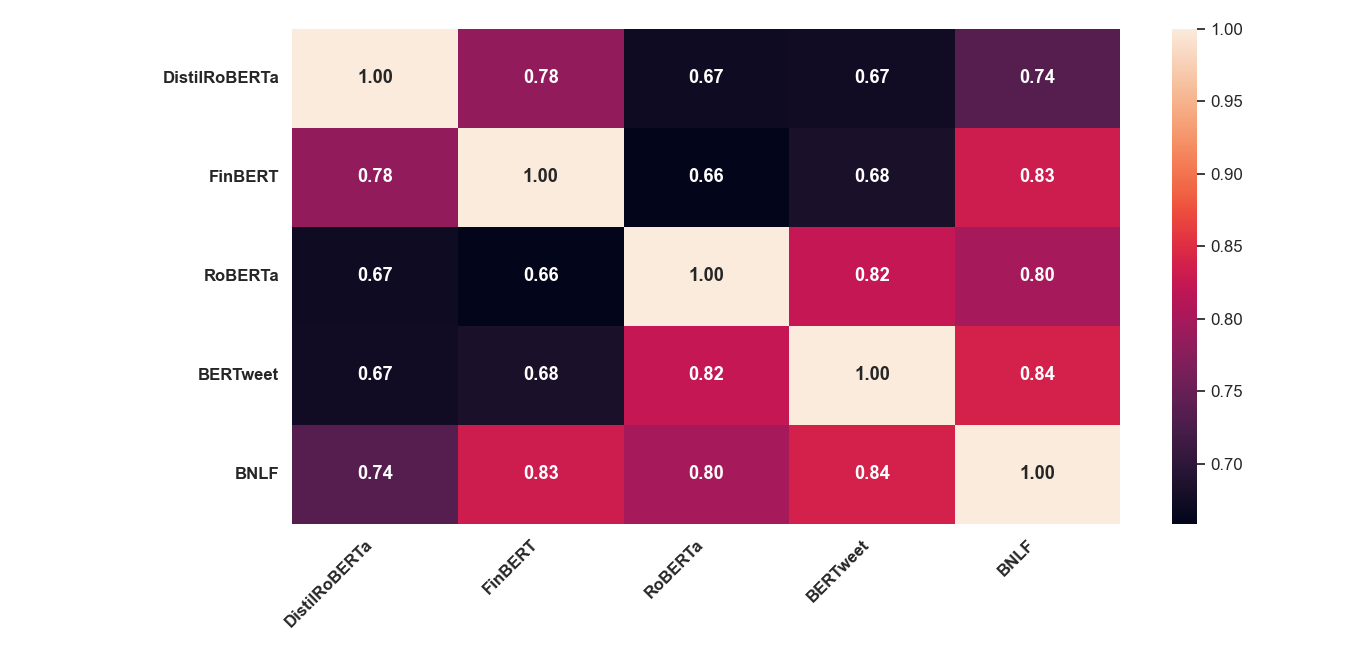}
\caption{Heatmap of pairwise agreement scores between individual LLMs and BNLF. 
Darker shades indicate stronger agreement, corresponding to higher proportions of matching sentiment labels.}
    \label{fig:agreement_heatmap}
\end{figure}

Cohen’s Kappa values, reported in Table~\ref{tab:agreement_combined}, adjust for chance agreement and provide a stricter measure of alignment.\footnote{Kappa values are generally lower than raw proportions because they account for agreement that could occur randomly.} BNLF achieves the highest mean consistency (0.801), reinforcing its role as a balanced integrator of individual model predictions, and DistilRoBERTa’s agreements are moderate across all models. The strongest alignment is between FinBERT and BNLF ($\kappa=0.704$), while BERTweet also shows substantial agreement with both RoBERTa ($\kappa=0.643$) and BNLF ($\kappa=0.677$).
These agreement values reflect differences in domain adaptation rather than inconsistency among models. For instance, the RoBERTa–DistilRoBERTa pair shows lower agreement, while BERTweet’s relatively high overall agreement suggests that it captures linguistic features shared across both formal and informal text sources. Furthermore, the difference between raw agreement and Kappa values indicates that the observed agreement patterns, whether high or low, are statistically meaningful and not due to chance.

\subsection{Analysing Sentiment Predictions of BNLF}
\label{bn_fusion}

In this section, we conduct additional analyses on the BNLF model, including inference behaviour and influence strength assessment. These evaluations aim to reveal how individual LLMs and dataset characteristics contribute to the BNLF’s final sentiment predictions, providing a clearer understanding of its  decision-making process, and supporting transparency and explicability, key aspects of trustworthy AI.

\subsubsection{ Inference Analysis}

 Inference analysis is a standard step in BN studies, as it reveals how the model updates beliefs under different evidence settings~\citep{amirzadeh2025dynamic}. We explore scenario-based inference to see how BNLF combines LLM predictions with contextual factors such as corpus type. In each scenario, we set specific nodes in the BN to fixed states and observe the resulting posterior probabilities for the sentiment output node. These controlled cases show whether the BN treats all model predictions equally or instead changes its output based on the characteristics of the dataset.

In the BN design, the \texttt{Input Text} is implemented as a deterministic node, since the dataset type is an observed fact rather than a probabilistic variable. Its inputs from various soucres (FinancialPhraseBank, {FIQA}, {TFNS}) are fixed for each experiment and conditions the behaviour of the LLM prediction nodes. This ensures that data-level effects are explicitly incorporated into inference without adding additional uncertainty at the dataset level.

\vspace{4pt}
\noindent\textbf{\textit{\small{Scenario 1 – Corpus Type Effect with Identical Model Predictions}}}\\

In the first scenario, we fix all LLM sentiment predictions to \textit{Negative} while changing the corpus type to investigate how the origin of the dataset influences the final sentiment distribution of the BNLF. Table~\ref{tab:bn_infer_corpus_allneg} presents the resulting posterior probabilities for each corpus setting.

\begin{table}[h!]
\centering
\caption{BNLF posteriors with all LLMs fixed to negative, under different corpus type settings. Values are percentages for negative, neutral, and positive sentiment.}
\label{tab:bn_infer_corpus_allneg}
\begin{tabular}{lccc}
\toprule
Corpus Type & Negative (\%) & Neutral (\%) & Positive (\%) \\
\midrule
Financial PhraseBank & 96.55 & 1.72  & 1.72 \\
FIQA                  & 95.06 & 3.70  & 1.23 \\
TFNS                  & 66.54 & 31.52 & 1.94 \\
\bottomrule
\end{tabular}
\end{table}

Even with identical negative predictions from all LLMs, BNLF’s outputs vary considerably by corpus. In particular, For {Financial PhraseBank} and {FIQA}, the outcome remains almost deterministic (over 95\% negative), whereas {TFNS} shifts toward greater uncertainty, with 66.5\% negative and 31.5\% neutral. This shows that corpus type can influence  the model’s certainty and also the balance between sentiment classes. This effect is further illustrated in Figure~\ref{fig:sen1} which visualises the {TFNS} case where {neutral} probability rises noticeably despite all models predicting {negative}.

\begin{figure}[h!]
\centering
\includegraphics[width=0.7\textwidth,keepaspectratio]{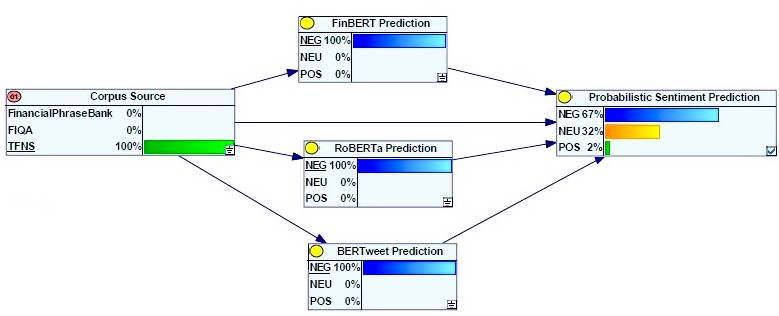}
\caption{BNLF inference for the {TFNS} corpus with all LLMs fixed to negative.  
The model outputs 67\% negative, 32\% neutral, and 2\% positive, showing a clear rise in neutral probability even though all models give negative as input.}
\label{fig:sen1}
\end{figure}

\vspace{9pt}
\noindent\textbf{\textit{\small{Scenario 2 – Corpus Type Variation with Model Disagreement}}}\\

In the second scenario, each LLM is set to a different sentiment state (FinBERT = negative, RoBERTa = neutral, BERTweet = positive) while varying the corpus type. This configuration creates maximum disagreement among the individual models, allowing us to examine how BNLF resolves conflicting evidence across different dataset domains. Table~\ref{tab:bn_infer_corpus_disagree} and Figure~\ref{fig:sen2} summarise the resulting posterior probabilities for BNLF under each corpus setting.

\begin{table}[h!]
\centering
\caption{BNLF posteriors with FinBERT = Negative, RoBERTa = Neutral, and BERTweet = Positive, under different corpus type settings. Values are percentages.}
\label{tab:bn_infer_corpus_disagree}
\begin{tabular}{lccc}
\toprule
Corpus Type & NEG (\%) & NEU (\%) & POS (\%) \\
\midrule
Financial PhraseBank & 66.67 & 16.67 & 16.67 \\
FIQA                  & 16.67 & 16.67 & 66.67 \\
TFNS                  & 4.55  & 36.36 & 59.09 \\
\bottomrule
\end{tabular}
\end{table}

Even with the disagreement pattern from all models, BNLF’s outputs change significantly with corpus type. For {Financial PhraseBank}, the prediction is mostly negative (66.7\%). For {FIQA}, positive becomes the dominant sentiment (66.7\%). For {TFNS}, positive remains highest (59.1\%), while neutral is also relatively high (36.4\%), showing greater uncertainty.  
This shows that corpus type can affect both the model’s certainty and which sentiment class becomes dominant when models disagree. As shown in Figure~\ref{fig:sen2}, the {FIQA} case illustrates that positive sentiment has the highest prediction probability despite mixed model inputs. 

\begin{figure}[h!]
\centering
\includegraphics[width=0.7\textwidth, keepaspectratio]{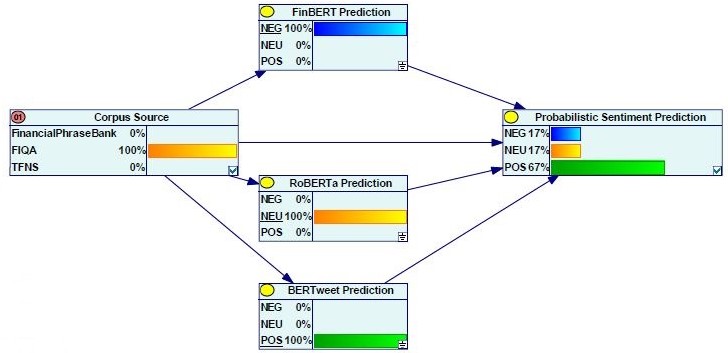}
\caption{BNLF inference for the {FIQA} corpus with FinBERT = Negative, RoBERTa = Neutral, and BERTweet = Positive. The model outputs 17\% negative, 17\% neutral, and 67\% positive, showing a clear shift toward positive sentiment despite conflicting inputs.}
\label{fig:sen2}
\end{figure}

\subsubsection{Influence Strength}
To further analyse the BNLF model, we explore the strength of influence, a measure of how strongly changes in a parent node affect the probability distribution of its child node. It is calculated from the CPTs as the average difference between the child’s probability distributions under different parent states.
The strength of influence results are shown in Figure~\ref{fig:influence}, where the arc annotations indicate influence values and thicker arcs represent stronger relationships between connected nodes.

\begin{figure}[h!]
\centering
\includegraphics[width=0.90\textwidth]{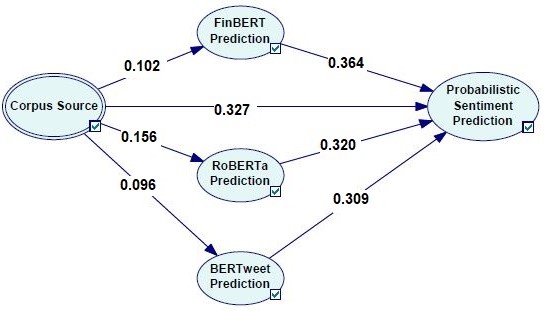}
\caption{Strength of influence diagram for BNLF. Arc thickness is proportional to the magnitude of influence between nodes.}
\label{fig:influence}
\end{figure}

The figure shows that FinBERT has the strongest direct influence on BNLF (0.364), followed by RoBERTa (0.320) and BERTweet (0.309). The corpus type also directly affects BNLF (0.327) and indirectly influences it through the LLMs, most notably RoBERTa (0.156), then FinBERT (0.102) and BERTweet (0.096). This indicates that, although all models contribute, FinBERT and RoBERTa are the main drivers of the final sentiment predictions, likely because FinBERT benefits from domain-specific financial training and RoBERTa offers strong general-purpose coverage. This aligns with earlier results showing that these two models have the largest effect on BNLF outputs under different corpus types.

The two inference scenarios and the influence analysis together demonstrate the benefits of BNLF in enhancing reasoning and interpretability. The results show that BNLF dynamically integrates varying model predictions and contextual factors, generating sentiment outcomes that reflect both consensus and uncertainty in a transparent way. This behaviour underscores its ability to combine evidence probabilistically, providing a clear understanding of how sentiment decisions are formed and revealing the mechanisms behind them. 

Beyond interpretability, these analyses also highlight BNLF’s advantage in causal reasoning.
A major limitation of current LLMs lies in their restricted capacity for causal reasoning, as they often struggle to capture underlying cause–effect dependencies within data, a critical gap in domains such as financial analysis, where understanding the drivers behind sentiments is essential for interpretability. For instance, \citet{jin2023cladder} show that LLMs still fail to reason consistently about causality, while \citet{liu2023trustworthy} demonstrate that GPT models can identify correlations between events and sentiments but often lack logical coherence in counterfactual reasoning. However, BNLF explicitly models directional dependencies, enabling inference from effect to cause and vice versa. As demonstrated by the inference and influence analyses, the Bayesian structure systematically identifies which LLM or contextual factor most plausibly contributes to a given outcome, even when model predictions disagree or appear counterintuitive. This causal interpretability enhances the framework’s transparency and directly supports the explicability principle of Trustworthy AI~\citep{thiebes2021trustworthy}.

\section{Conclusion and Future Work }\label{Conclusions}

This study addresses key challenges in applying large language models for financial sentiment analysis across formal financial data and social media posts. While LLMs have achieved strong performance, they often require costly fine-tuning, extensive prompt engineering, and still produce inconsistent results across domains and corpora. To overcome these limitations, we propose the Bayesian network LLM fusion framework, which integrates sentiment outputs from multiple LLMs. The framework shows that different LLMs contribute varying influences to sentiment prediction, and their probabilistic combination enhances overall performance.


Experimental results in three diverse, human-annotated financial datasets show that BNLF generally outperforms all individual LLM baselines. Moreover, it improves accuracy and weighted F1-score by around six percentage  over the baseline LLM, with gains observed across all datasets, including the challenging {FIQA} corpus. Class-level analysis shows that BNLF achieves balanced performance. Furthermore, the inference and influence-strength analyses reveal that FinBERT and RoBERTa exert the strongest influence, with BERTweet providing complementary signals. The source corpus of each text also plays an important role in shaping the certainty and sentiment distributions.  

Beyond quantitative gains, BNLF offers broader design advantages for developing reliable and adaptable LLM-based systems. It addresses two main challenges in this domain, including limited interpretability and scalable integration, by incorporating causal and probabilistic reasoning within a modular architecture. This design allows the framework to integrate heterogeneous models while maintaining transparency in its decision process.


While scalability in LLMs is traditionally defined by increasing model parameters, data size, and computational resources~\citep{xiong2024temporal, amjad2025review}, BNLF achieves both structural and functional scalability, where adaptability is achieved without retraining large models. It allows the seamless addition or replacement of LLMs~ (structural scalability) and, more importantly, supports adaptation to different tasks~(functional scalability). This modular architecture enables the framework to adapt or extend dynamically, representing a shift from parametric scaling to architectural scalability. Building on this scalable design, BNLF offers a flexible foundation for several promising research directions aimed at enhancing its scalability, interpretability, and applicability across broader financial and generative contexts.

In line with this scalable design, future research could extend BNLF along two complementary directions, addressing some of the current framework’s limitations. First, structural scalability can be enhanced by incorporating additional LLM architectures or multilingual financial corpora to strengthen cross-domain robustness and adaptability. Moreover, constructing the BN directly from the confidence scores of sentiment prediction, rather than discrete sentiment classifications, using structure-learning algorithms such as the PC algorithm, could reveal more nuanced dependencies among model outputs. Second, functional scalability involves expanding the framework beyond sentiment classification. A promising direction is to extend BNLF to generative modelling, where multiple GPT-style models contribute next-token probabilities and the BN fuses them to select the most likely output. Another direction is to employ dynamic BNs to capture temporal sentiment evolution and its influence on financial markets~\citep{amirzadeh2025dynamicdbn}.

\section*{Data availability}
The datasets used in this study were obtained from publicly available resources on Hugging Face. 
The Bayesian network model is available at: \url{https://doi.org/10.59381/ceglcyvwla}

\appendix

\section{Supplementary Results}\label{sup_res}
In addition to the models discussed in the main text, we evaluated other widely used and trending sentiment models from Hugging Face, including {ALSGYU General} and {TabSA Multilingual}. Despite their popularity and accessibility, their performance on the combined dataset was notably weaker, as shown in Table~\ref{tab:appendix_baselines}.
\begin{table}[h!]
\centering
\caption{Performance of additional baseline models on the combined test set. Best results among these supplementary models are shown in bold.}
\begin{tabular}{lccc}
\hline
\textbf{Model} & \textbf{Accuracy} & \textbf{Macro-F1} & \textbf{Weighted-F1} \\
\hline
ALSGYU General     & \textbf{0.452} & \textbf{0.312} & \textbf{0.434} \\
TabSA Multilingual & 0.302 & 0.299 & 0.293 \\
\hline
\end{tabular}
\label{tab:appendix_baselines}
\end{table}


\bibliographystyle{elsarticle-harv}

\end{document}